\documentclass{article}

\usepackage{microtype}
\usepackage{graphicx}
\usepackage{subfigure}
\usepackage{booktabs} 

\usepackage{hyperref}
\usepackage{color,soul}

\usepackage[accepted]{icml2019}

\icmltitlerunning{E-Sports Scouting Based on Twitch}

\begin{document}

\twocolumn[
\icmltitle{E-Sports Talent Scouting Based on Multimodal Twitch Stream Data}



\icmlsetsymbol{equal}{*}

\begin{icmlauthorlist}
\icmlauthor{Anna Belova}{equal,cmu}
\icmlauthor{Wen He}{equal,cmu}
\icmlauthor{Ziyi Zhong}{equal,cmu}
\end{icmlauthorlist}

\icmlaffiliation{cmu}{Language Technologies Institute, Carnegie Mellon University, Pittsburgh, PA}

\icmlcorrespondingauthor{Anna Belova}{abelova@alumni.cmu.edu}

\icmlkeywords{e-sports, Twitch stream, multimodal learning, Bayesian pooling}

\vskip 0.3in
]



\printAffiliationsAndNotice{\icmlEqualContribution} 

\begin{abstract}
  We propose and investigate feasibility of a novel task that consists in finding e-sports talent using multimodal Twitch chat and video stream data. In that, we focus on predicting the ranks of Counter-Strike: Global Offensive (CS:GO) gamers who broadcast their games on Twitch. During January 2019--April 2019, we have built two Twitch stream collections: One for 425 publicly ranked CS:GO gamers and one for 9,928 unranked CS:GO gamers. We extract neural features from video, audio and text chat data and estimate modality-specific probabilities for a gamer to be top-ranked during the data collection time-frame. A hierarchical Bayesian model is then used to pool the evidence across modalities and generate estimates of intrinsic skill for each gamer. Our modeling is validated through correlating the intrinsic skill predictions with May 2019 ranks of the publicly profiled gamers.
\end{abstract}

\section{Introduction}
E-sports are formalized competitive computer gaming \citep{taylor2018watch}. E-sports are gaining attention similar to that of the real life sports for their entertainment value \citep{raath2017rise}. As a consequence, e-sports talent scouting is likely to become an economically valuable activity.

\begin{figure}[h]
    \centering
    \includegraphics[width=.45\textwidth]{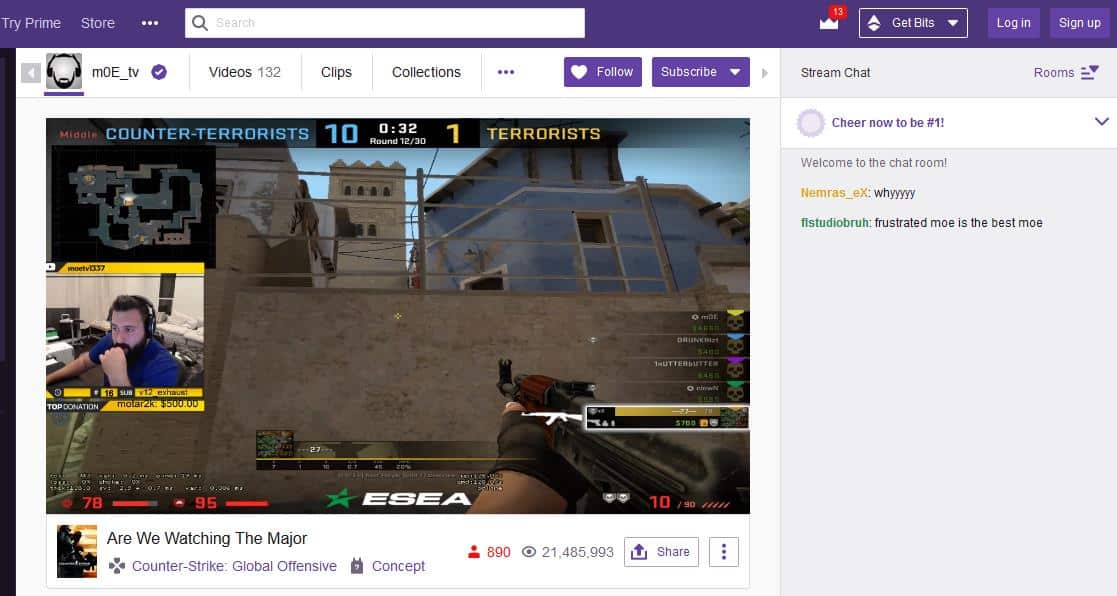}
    \caption{Example of the Video Stream and Chat}
    \label{fig:example}
\end{figure}

Many gamers share their real-time online game experience on Twitch,\footnote{\href{https://www.twitch.tv}{https://www.twitch.tv}} which is an interactive video streaming platform. The video component of the broadcast typically contains game visuals, but could also include an image of a gamer him or herself (see Figure~\ref{fig:example}). The audio component of the broadcast could contain both, the sounds of the game as well as the commentary of the gamer. In turn, the broadcast audience can provide real time feedback on the game being played via the text chat. Twitch is now the fourth largest source of peak Internet traffic in the US \citep{deng2015behind}.

We propose a novel multimodal task that consists in automated identification of talented online gamers based on their Twitch streamed video and audio recordings as well as their spectator chat messages. To this end, we focus on the Counter-Strike: Global Offensive (CS:GO),\footnote{CS:GO is a multi-player first-person shooter video game developed by Hidden Path Entertainment and Valve Corporation. It is one of the most popular games represented on Twitch.} for which some of the gamers are publicly profiled by the E-Sports Entertainment Association (ESEA) League.\footnote{\href{https://play.esea.net}{https://play.esea.net}} 

Our research contributions are two-fold. First, we develop a high recall baseline model demonstrating feasibility of the e-sports talent scouting task outlined above. Second, we contribute two new multi-modal Twitch chat and video stream datasets (one for 425 publicly ranked CS:GO gamers and one for 9,928 unranked CS:GO gamers) collected during January 2019--April 2019.\footnote{The data acquisition and modeling code can be accessed at \href{https://github.com/mug31416/E-sports-on-Twitch}{https://github.com/mug31416/E-sports-on-Twitch}. We are working on making the data publicly available.}

\section{Related Work}

Although multimodal research on e-sports is rather scarce, we found real-life sports research and research using Twitch platform data to be pertinent to our proposed task. 

\subsection{Research Related to Sports Talent Scouting}
There is considerable literature focusing on the real life competitive sports performance analysis. In that, there are a few examples of multimodal research that has focused on mining for athletic talent using a variety of sources, including biomedical data and social media statistics \citep{schumaker2010sports,radicchi2016social,cumming2017bio}. 

More broadly, there are several extensively studied tasks that can be seen as pretext tasks for multimodal sports performance analysis. These tasks include multimodal sentiment detection \citep{cai2015convolutional,borth2013sentibank,campos2017pixels,thelwall2012sentiment,he2018multimedia},  event detection \citep{ma2018joint,aslam2018towards,ma2018tree,wang2018eann}, and clustering \citep{wang2007learning}. Tools and methods employed in this literature on intelligent multimedia modeling, including the use of neural features \citep{kung1998neural} as well as work with incomplete and heterogeneous data \citep{miech2018learning}, are directly relevant to our task.

\subsection{Research using Twitch Platform Data} 
Relevant research employing Twitch data has largely focused on chat \citep{ruvalcaba2018women,ford2017chat,nakandala2017gendered} or mixed-mode (chat and audio transcripts) \citep{wirman2018discourse} communication as well as on retrieval of streaming video highlights \citep{zhang2017seeker,liu2015live}, with some studies employing multimodal information \citep{jianglightor,fu2017video}. Thus far, game play on Twitch explored a problem of identifying trolls in Pokemon games \citep{haque2019twitch}. 

There are also many studies on Twitch as a social and economic phenomenon \citep{deng2015behind,taylor2018watch,zhu2017understanding,gros2018interactions,sjoblom2019ingredients}, with papers ranging from qualitative exploration to quantitative analysis of platform participation activity and user surveys. However,  optimization of Twitch content delivery thus far appears to have received most attention \citep{glickman2018design,provensi2017cloud,bilal2017impact,laterman2017campus,deng2018profiling}. 

To our knowledge, there is no research that combines Twitch chat and video stream data with an external supervision signal from a public gaming leaderboard to make inferences about comparative player performance.

\section{Data}\label{sec:data}
ESEA and Twitch data for modeling purposes have been systematically collected during January 2019--April 2019. Upon completion of modeling, we have obtained \textit{ad hoc} validation data from May 2019 ESEA profiles.\footnote{Because of the changes in ESEA website security policy, the May 2019 rankings could not be obtained using the same scraping scripts as the ones used for the main data acquisition. Hence, we relied on a one-at-a-time manual rank look-up for selected gamers.} We provide data collection details below. 

\subsection{ESEA data}
The ESEA data comprise the supervision element of our study. The ESEA League is an online e-sports community that matches CS:GO players across different skill levels, maintains player statistics, and issues performance-based ranks. The ESEA leaderboard contains a ranked list of approximately 2,000 players who fall into four ranking sections---A (highest), B, C, and D---with each category containing approximately 500 players. The leaderboard rank sections are updated monthly. In addition, there are rank sections G (pre-professional) and S (professional). Elite gamers in these sections are tracked separately from the public leaderboard; there are approximately 500 players in this group.

Some ESEA player profiles contain Twitch stream identifiers. On two separate occasions (in January 2019 and in February 2019), we have scraped the ESEA leaderboard to obtain a roster of CS:GO players for tracking on Twitch. As a result, we extracted 748 CS:GO gamers with Twitch accounts for monitoring. 

In addition, for each monitored gamer we extracted his or her rank section in January 2019, February 2019, March 2019, and April 2019. Because of the substantial leaderboard turnover, not all gamers have rank data each month (some get promoted to the elite sections, while others withdraw from competitive play). We also note that our monitored group has a higher share of rank A section players (40\% vs 25\% in general).

\subsection{Twitch data}
Twitch streamed video, audio, and text chat for CS:GO games comprise the feature space of our study. As explained in Section~\ref{sec:methods}, we also employ the number of Twitch user followers as a weak supervision signal for pretraining.

The Twitch platform has a well-developed API\footnote{\href{https://dev.twitch.tv/docs}{https://dev.twitch.tv/docs}} that can be used to detect specific streamer status (i.e., whether the streamer is online), extract all channels that are streaming CS:GO in general, and acquire channel-specific chat logs. Furthermore, Twitch provides an API, together with ``streamlink'' that allows us to record live streams with audio.  

We have continuously collected Twitch chat and stream data during January 2019--April 2019 for the 748 ESEA players with available Twitch stream identifiers.  In addition, for use in pretraining, we have also collected these data for all CS:GO games during March 2019--April 2019. 

\subsubsection{Followers}
 Meta-data for each Twitch user contains a number of followers (i.e. other users who are subscribed to this user's channel). Twitch provides an easy API to acquire number of followers of a channel. We obtained an April 2019 snapshot of the number of followers for each targeted Twitch user. Note that for 2\% of the Twitch users we tracked we have been unable to obtain the number of followers due to various exceptions (e.g., Twitch account anomaly, account deleted).

\subsubsection{Video Stream}
We have followed the guidelines in this tutorial,\footnote{\href{https://www.godo.dev/tutorials/python-record-twitch/}{https://www.godo.dev/tutorials/python-record-twitch/}} to set up continuous monitoring for several groups of targeted Twitch users and capture the video stream whenever one of the users in the group appears online. For the unranked CS:GO players, we have set up a query service to obtain a list of users playing CS:GO every 15 seconds; this list has been passed on to the video stream recorder.

We have run multiple instances of this streamer monitor in parallel. Because we only record one stream at a time (within a group of users), it is possible that we have missed some streams. However, we do not see this as a major issue, because not many players are be online at the same time and we record the stream for a limited amount of time. Specifically, we have limited the stream recording duration to 5 minutes. We also have limited the number of clips per user per day to be one. 

For the 748 ESEA streamers, we have collected 3,330 videos by 368 distinct gamers. The total size of these data is 741 GB. For the all other CS:GO gamers on Twitch without the ESEA profile, have collected 13,969 videos by 9,928 distinct users. The size of these data is 1.56 TB.

\subsubsection{Text Chat}
To gather the chat data from Twitch streams, we followed the pipeline in this GitHub repository.\footnote{\href{https://github.com/bernardopires/twitch-chat-logger}{https://github.com/bernardopires/twitch-chat-logger}} This pipeline automatically creates bots to listen for the chat log data of the target players. One bot is responsible for 20 channels. We have monitored 748 ESEA target gamers, which has required 38 bots in total.

For general CS:GO collection, we first used Twitch API to scrape live streams of CS:GO game and added the limitation of stream language to be English. After we got the names of those live streams, we used the same way of collecting chat logs for target gamers but fed in those live stream names as the monitor targets. Since the live streams are only active for a limited time period, we repeated the above data collection process once a day.

For the 748 ESEA streamers, there are 210,185 chat logs collected from 308 distinct players (channels). The total size of these data is 47 MB. For the general CS:GO gamers on Twitch without the ESEA profile, we have collected 1,020,687 chat logs from 660 distinct players. The size of those general CS:GO gamers is 187 MB.

\begin{table}[h]
    \centering
    \caption{Modeling Data Statistics (Validation / Training)}
    \label{tab:data_stats}
    \begin{tabular}{cccc}
         \toprule
         \textbf{Modality} & \textbf{Task} & \textbf{\#Users} & \textbf{\#Datapoints}  \\
         \midrule
         Video & pretrain & 186/ 3,531 & 433/ 8,221 \\
         Video & fine-tune & 98/ 270 & 809/ 2,521 \\
         Audio & pretrain & 1,822/ 4,094 & 1,994/ 5,013 \\
         Audio & fine-tune & 91/ 261 & 613/ 1,727 \\
         Chat & pretrain & 464/ 539 & 3,825/ 11,473\\
         Chat & fine-tune & 79/ 203 & 2,421/ 6,358\\
         \bottomrule
    \end{tabular}
\end{table}


\section{Methods}\label{sec:methods}

\begin{figure}[h]
    \centering
    \includegraphics[width=.45\textwidth]{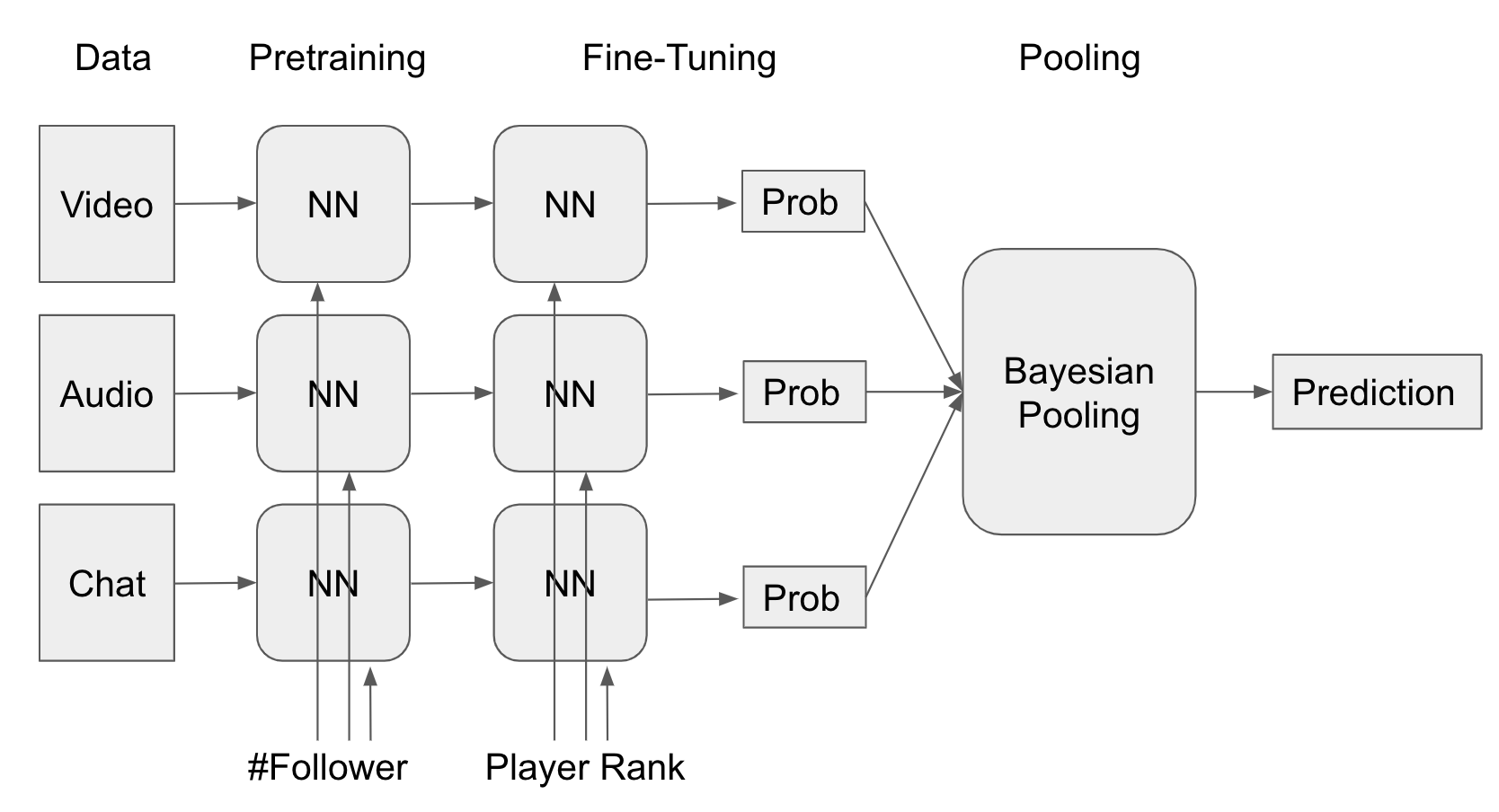}
    \caption{Modeling Pipeline Overview}
    \label{fig:overview}
\end{figure}

Figure~\ref{fig:overview} provides an overview of our modeling pipeline. In that, we first apply discriminative learning techniques separately for each modality with an aim to predict whether or not a gamer is in the rank A section of the ESEA leaderboard at any point during January 2019--April 2019 (i.e., a binary classification task).\footnote{We did not use month-specific gamer ranks due to the data alignment challenges.} To this end, we use a training set of the ESEA Twitch streamers, which, depending on the modality-specific data availability, comprises approximately 75\% of all ESEA Twitch streamers.\footnote{The split of the ESEA Twitch streamers has been implemented to preserve the share of the rank A section gamers of 40\% across the training set and validation set.} We experiment with two discriminative learning strategies, one using \textit{cross-entropy loss} and another using \textit{triplet loss} \citep{cai2007spectral,hermans2017defense}. The setting of triplet loss is \texttt{maxplus} with square euclidean distance. The generated embeddings were then fed into an SVM of \texttt{scikit-learn} default settings; probabilistic predictions are generated using the Platt method \citep{platt1999probabilistic}.\footnote{Our preliminary evaluation has shown that a non-linear SVM is an effective classifier for our data. It performed better than logistic regression, and random forest classifier.}

As can be seen from Table~\ref{tab:data_stats}, the number of unique Twitch streamers with available training data ranges between 203 and 270. Therefore, in addition to training our models from scratch, we experiment with \textit{fine-tuning} these models after  \textit{pretraining} them on a related task of predicting the quantized number of Twitch followers.\footnote{There is a statistically significant 8\% correlation between rank A section membership and the number of followers.} The quantization has been done according to the deciles of the empirical distribution for the number of followers; we have treated this pretraining target as a multiclass prediction problem.

In the second modeling stage, we use a hierarchical Bayesian model to pool the evidence from the upstream modeling tasks (i.e., predicted probability for a gamer to be in the rank A section) as well as the validation set rank section data. This modeling stage generates estimates of the intrinsic skill level for each gamer in our dataset. Below we provide additional details on the first-stage modality-specific data processing and modeling, as well as on the second-stage pooling model. 

\subsection{Visual Data}
For the visual data we have used a classic two-stream modeling approach \citep{simonyan2014two}, in which spatial/image data and motion/optical flow data are modeled using two separate networks.

\subsubsection{Processing}
We have used \texttt{ffmpeg} \citep{FFMPEG2010} to cut all ESEA gamer videos to 60 sec, whereas videos used for pretraining have been cut to 20 sec. All videos have been also downsampled to 10 frames per second. \texttt{OpenCV} \citep{opencv_library} has been used to rescale all frames such that the smallest dimension is 256 pixels; the resulting JPEG quality has been set at 60\%.\footnote{We note that performance of our models for JPEG quality above 60\% has not been materially better than performance reported in this paper.} Optical flow has been estimated using a variational motion estimation method \citep{zach2007duality} from a GPU adaptation of \texttt{OpenCV} functionality from this GitHub repository.\footnote{\href{https://github.com/feichtenhofer/gpu\_flow}{https://github.com/feichtenhofer/gpu\_flow}}

\subsubsection{Modeling}
We adapted the two-stream network implementation from this GitHub repository.\footnote{\href{https://github.com/jeffreyhuang1/two-stream-action-recognition}{https://github.com/jeffreyhuang1/two-stream-action-recognition}} For both, spatial network and motion network, we trained the projection layer and the last block of the pretrained ResNet101 \citep{he2016deep}, available from \texttt{torchvision} \citep{Marcel:2010:TMP:1873951.1874254}. We use initial learning rates of 1e-3 and 5e-3 for the spatial model and the motion model, respectively. Each model has been fit for 10 epochs, using batch size four. 

\subsection{Audio Data}
While we experimented with both, Mel-frequency cepstral coefficients (MFCC) and transcripts, we determined that audio transcripts are not usable because around 25\% of Twitch audio streams are non-English. Hence, we focused only on MFCC features.

\subsubsection{Processing}

We have extracted of audio (WAV format) from videos (MP4) using \texttt{ffmpeg} \citep{FFMPEG2010}. And extraction of MFCC features from WAV files using \texttt{librosa}. \footnote{\href{https://librosa.github.io/}{https://librosa.github.io/}} MFCC latent dimension is set to 20 and the audio is collected at a sample rate of 8K.


\subsubsection{Modeling}

Instead of clustering MFCCs into a codebook, we have trained a Convolutional Neural Network (CNN) architecture directly on MFCCs using \texttt{keras}.\footnote{\href{https://keras.io/}{https://keras.io/}}  Three region sizes (2, 3, 4) are used, corresponding to bi-gram, tri-gram and 4-gram and for each region size, 16 filters. After a Global Max Pooling, there is a fully connected part with one hidden layer of 50 dimensions. 

\subsection{Chat Data}
To represent information contained in the Twitch chat logs, we developed two models, one focusing on the chat content and another focusing on the chat density received.

\subsubsection{Processing} 

For the first model, we grouped the chat messages based on the time-stamps: Messages sent within a certain time period (one hour in our case) are grouped together as one feature element. We first trained a model using \texttt{fastText} library \footnote{\href{https://github.com/facebookresearch/fastText/tree/master/python}{https://github.com/facebookresearch/fastText/tree/master/python}} using the \texttt{train\_unsupervised} method provided with model set to \texttt{skipgram} and dimension set to 500. We then used the trained model to create sentence-level embeddings for every chat log. Embeddings for chat logs within one group were stacked together to form one feature point.

For the second model, we represented the temporal information of chat as a vector of chat logs received by each player per day. For each player, we counted the number of comments received every day and created a vector recording those numbers. The lengths of the chat density feature vectors are variable for each player since the coverage of stream days is different.

\begin{figure}[h]
    \centering
    \includegraphics[width=.40\textwidth]{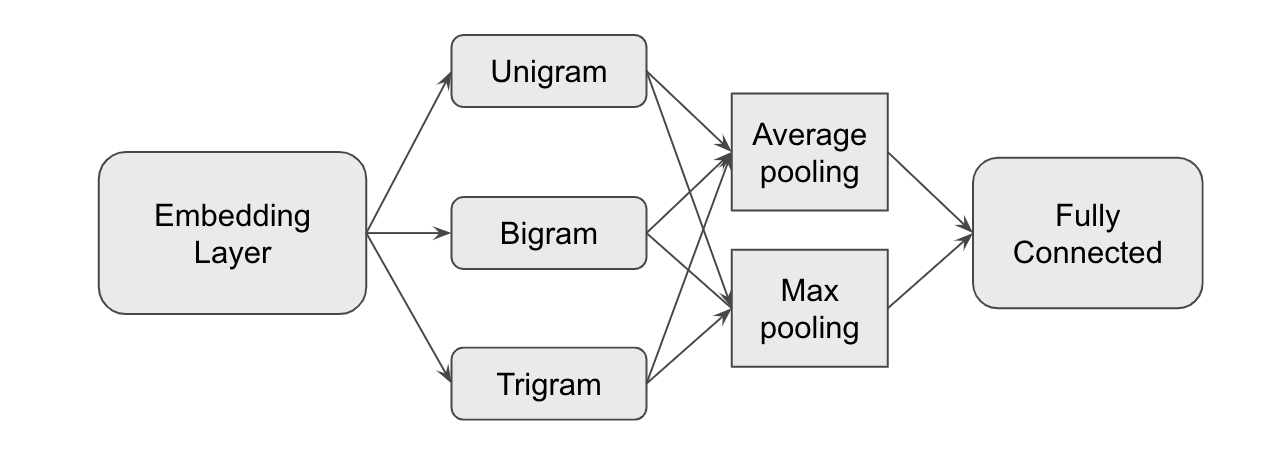}
    \caption{Chat Modeling Architecture}
    \label{fig:chat_model}
\end{figure}

\subsubsection{Modeling} 

In both pretraining and fine-tuning stage, the language classifier is an embedding layer followed by a concatenation of the unigram, bigram and trigram extractor with average pooling and max pooling. The model architecture is depicted in Figure \ref{fig:chat_model}. We use the \texttt{fastText} sentence embedding as the pretrained weight for the embedding layer. Since the chat density feature is also sequential, the same architecture is applied except that the input is the number of comments received instead of the actual sentence tokens.
The number of hidden units for all three n-gram extractors is set to be 256. A dropout rate of 0.3 is applied before the final linear layer. The model uses 1 as the batch size to avoid feature padding issues. The learning rate is 1e-3 throughout the training process. 

\subsection{Pooling}
The second modeling stage has focused on synthesis of the rank A section probability estimates from the upstream models. This stage has involved three modeling challenges. First, each upstream model has generated multiple predictions per gamer.\footnote{For example, the visual data models produce video-level predictions, whereas the text chat model produces chat cluster-level predictions.} Second, not all gamers have data for each modality and, hence, the corresponding upstream modeling result. Third, the validation set of ranked gamers is limited to 79--98 gamers, implying sampling uncertainty concerns.

\begin{figure}[h]
    \centering
    \includegraphics[width=.40\textwidth]{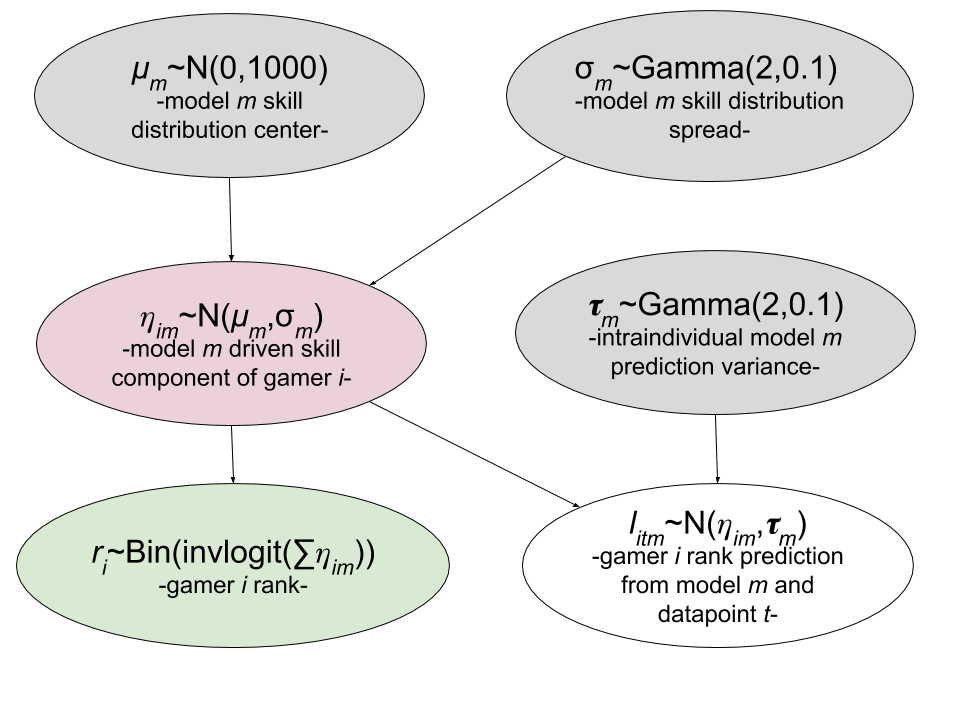}
    \caption{Bayesian Pooling Model}
    \label{fig:pooling_model}
\end{figure}

The pooling challenges above have been addressed using a generative modeling approach. Specifically, we have implemented a hierarchical Bayesian pooling model summarized in Figure~\ref{fig:pooling_model}. In this model, we assumed that each gamer's intrinsic/latent skill is a sum of modality-specific random effects $\eta_{im}$ drawn from a normal distribution with hyper-parameters $\mu_m$ (the grand mean) and $\sigma_m$ (the grand variance). To ensure that we obtain a meaningful representation of latent skill, we also assumed that gamer $i$-specific observed rank A status $r_i$ as well as logits $l_{itm}$ from each upstream model $m$ for gamer's modality-specific data point $t$ are generated by a distribution whose mean is centered at $\eta_{im}$. In addition, we allow for model $m$-specific source of variability $\tau_m$ to affect $l_{itm}$. Finally, to avoid using the data twice, the observed ranks A standings of the gamers in the training set used for upstream modeling have not been included in this model (as such, for these gamers the evidence of skill is represented only by the logit estimates). Additional details on the priors for each hyper-parameter are given in Figure~\ref{fig:pooling_model}. 

The samples from the posterior distribution for $\eta_i$ and other model parameters have been generated via Markov Chain Monte Carlo (MCMC) sampling using NUTS implemented in Stan \citep{carpenter2017stan}. We took 20,000 samples from four independent chains. Convergence has been confirmed on the last 10,000 draws using the Rhat statistic \citep{gelman1992inference}. We represent the posterior distribution for $\eta_i$ using 40,000 samples (10,000 warm-up samples on four chains).

\section{Evaluation}
For the intrinsic evaluation of our modeling we use standard metrics such as the area under the ROC curve (AUC), F1, precision, and recall. These metrics are evaluated on the validation set of the ESEA gamers, using gamer-level averaging of the modality-specific scores. The threshold for F1, precision, and recall has been selected to optimize the F1 measure on the training set of ESEA gamers (also using scores averaged at the gamer-level). Finally, we use the maximum \textit{a posteriori} probability (MAP) measure to represent distributional intrinsic skill estimates for the evaluation. 
    
Extrinsic evaluation of our modeling consists in validating our findings using the data from the May 2019 ESEA ranks. This evaluation is not exhaustive because of the restrictions placed by ESEA on the number of gamer profile queries. Specifically, we have checked three groups of gamers: (1) gamers in the top 10\% of the estimated skill distribution (42 gamers); (2) gamers in the bottom 10\% of the estimated skill distribution (42 gamers); (3) gamers for whom we estimated high intrinsic skill, yet whose rank section was below A (63 gamers). We have compared the preponderance of rank A or higher in each group. We also statistically tested the association between rank improvements for the first and the third group using Kendall's test of rank correlation \citep{kendall1938new}. 

\section{Results}\label{sec:results}

Table~\ref{tab:rank_prediction_result} shows the intrinsic evaluation results of our modeling on the validation set of ESEA gamers. In all cases, our target is to predict whether a player is in the rank A section. Models using the triplet loss have produced  gamer-specific rank A section probability scores that are better than random (i.e., AUC above 0.5) for all feature sets. On the other hand, models using the cross-entropy loss often generated worse than random predictions on the validation set. Pretraining on a related task has been helpful for models using audio features, chat text features, and chat temporal features. However, models using video features have not been improved by pretraining. 

Table~\ref{tab:rank_prediction_result} also shows that, across modalities, the chat temporal features have produced the highest AUC of 0.784, F1 score of 0.709, and precision of 0.683. The best recall of 0.925 has been produced by the video motion features with cross-entropy loss.\footnote{A model using cross-entropy loss with pretraining on the video spatial features has generated an even higher recall of 0.950. Yet, this model has also produced a worse than random AUC. Hence, this recall result needs to be interpreted with caution.}
Finally, while audio features appear to capture some of the association with the rank A section (AUC of 0.585), they are never the best in any of the other evaluation categories (i.e., F1, precision, and recall). 

The pooled model used the best performing model for each feature set. It generated the best overall AUC of 0.797 as well as the highest F1 of 0.754 and the second highest recall of 0.86 and precision of 0.672. Note, however, that the results of the Bayesian pooling are not directly comparable to the modality-specific results for two reasons. First, in contract to the upstream models that generate MLE estimates, this model produces a posterior density of gamer-specific skill and we use MAP to summarize it. Second, unlike the upstream models, the pooling model is estimated using the rank information of the validation set.

\begin{table}[h]
\centering
   \caption{Results for Validation Set Gamers.}
    \label{tab:rank_prediction_result}
\begin{tabular}{@{}llllll@{}}
\toprule
\textbf{Loss}   & \textbf{Pretr.}   & \textbf{AUC}   & \textbf{F1}    & \textbf{Prec.} & \textbf{Recall} \\ \midrule
\multicolumn{6}{c}{Audio Features}                                                                                \\\midrule
Cross-entropy   & yes                    & 0.339          & n.a.           & n.a.               & n.a.            \\
Cross-entropy   & no                     & 0.473          & n.a.           & n.a.               & n.a.            \\
Triplet+SVM     & yes                    & \textbf{0.585} & \textbf{0.526} & \textbf{0.417}     & \textbf{0.714}  \\
Triplet+SVM     & no                     & 0.527          & 0.411          & 0.395              & 0.429           \\\midrule
\multicolumn{6}{c}{Text Chat Features}                                                                            \\\midrule
Cross-entropy   & yes                    & 0.482          & 0.539          & 0.471              & 0.632           \\
Cross-entropy   & no                     & 0.508          & 0.506          & 0.467              & 0.553           \\
Triplet+SVM     & yes                    & \textbf{0.551} & \textbf{0.600} & 0.484              & \textbf{0.789}  \\
Triplet+SVM     & no                     & 0.531          & 0.558          & \textbf{0.500}     & 0.632           \\\midrule
\multicolumn{6}{c}{Temporal Chat Features}                                                                        \\\midrule
Cross-entropy   & yes                    & 0.576          & 0.564          & 0.550              & 0.579           \\
Cross-entropy   & no                     & 0.553          & 0.549          & 0.472              & 0.658           \\
Triplet+SVM     & yes                    & \textbf{0.784} & \textbf{0.709} & \textbf{0.683}     & \textbf{0.737}  \\
Triplet+SVM     & no                     & 0.625          & 0.550          & 0.524              & 0.579           \\\midrule
\multicolumn{6}{c}{Spatial Video Features}                                                                        \\\midrule
Cross-entropy   & yes                    & 0.489          & \textbf{0.585} & 0.422              & 0.950           \\
Cross-entropy   & no                     & 0.496          & 0.574          & 0.416              & 0.920           \\
Triplet+SVM     & yes                    & \textbf{0.609} & 0.554          & 0.459              & 0.700           \\
Triplet+SVM     & no                     & 0.606          & 0.571          & \textbf{0.462}     & \textbf{0.750}  \\\midrule
\multicolumn{6}{c}{Motion Video Features}                                                                         \\\midrule
Cross-entropy   & yes                    & 0.557          & 0.597          & 0.440              & \textbf{0.925}  \\
Cross-entropy   & no                     & 0.588          & \textbf{0.607} & 0.451              & \textbf{0.925}  \\
Triplet+SVM     & yes                    & 0.586          & 0.536          & 0.456              & 0.650           \\
Triplet+SVM     & no                     & \textbf{0.708} & 0.603          & \textbf{0.461}     & 0.875           \\ \midrule
\multicolumn{6}{c}{Bayesian Pooling*}                                                                              \\ \midrule
\multicolumn{2}{l}{Best upstream models} &   0.797 &	0.754 &	0.672 &	0.860            \\ \bottomrule
\end{tabular}
\raggedright Note: *Pooled results are MAP. n.a. -- the model generated to the majority class prediction, no optimal threshold could be estimated.
\end{table}

\begin{figure}[h]
    \centering
    \includegraphics[width=.45\textwidth]{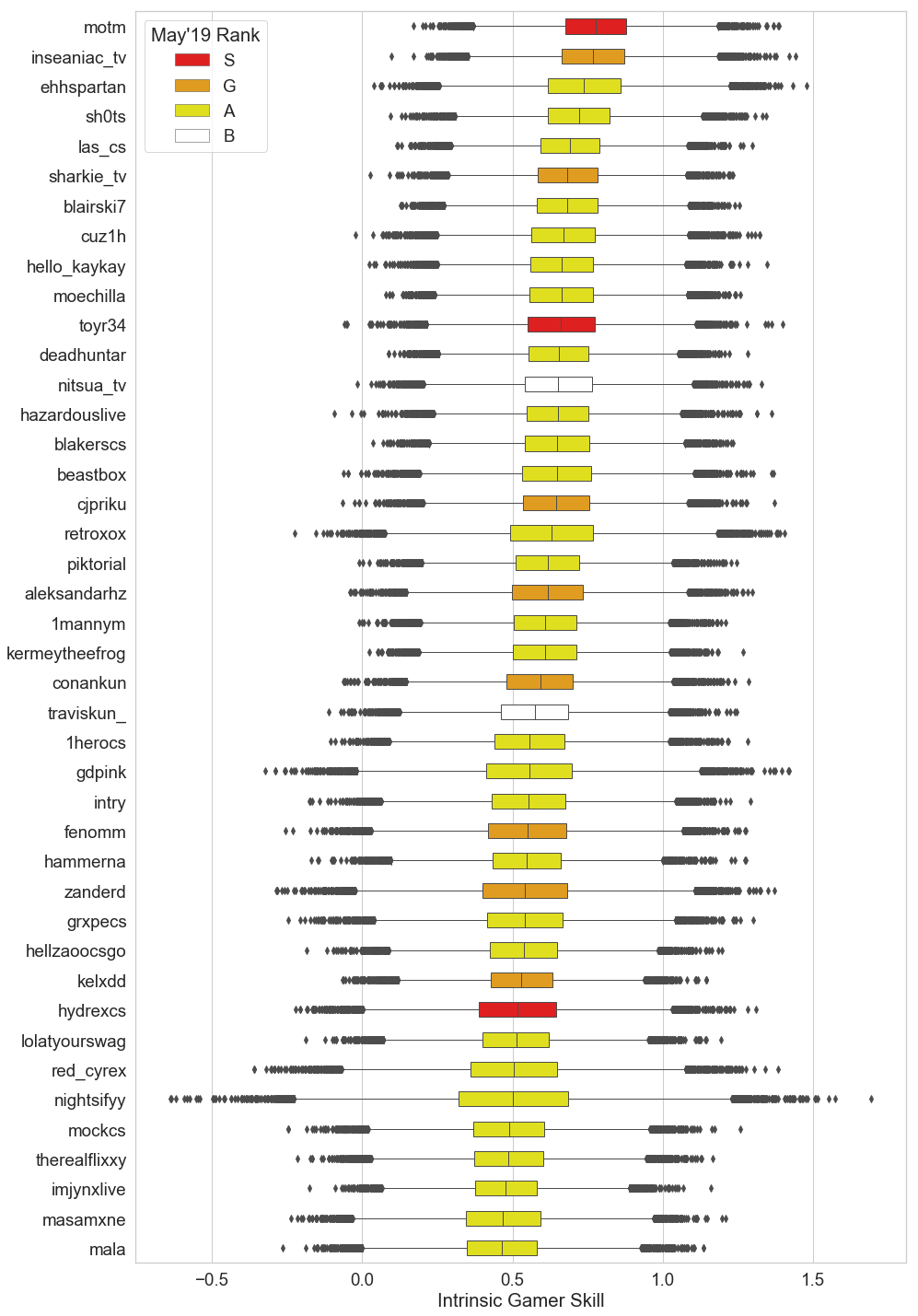}
    \caption{Gamers with the Top 10\% Jan'19-Apr'19 Data-based Intrinsic Skill Estimates and Their May'19 Ranks.}
    \label{fig:top_players}
\end{figure}
\begin{figure}[h]
    \centering
    \includegraphics[width=.45\textwidth]{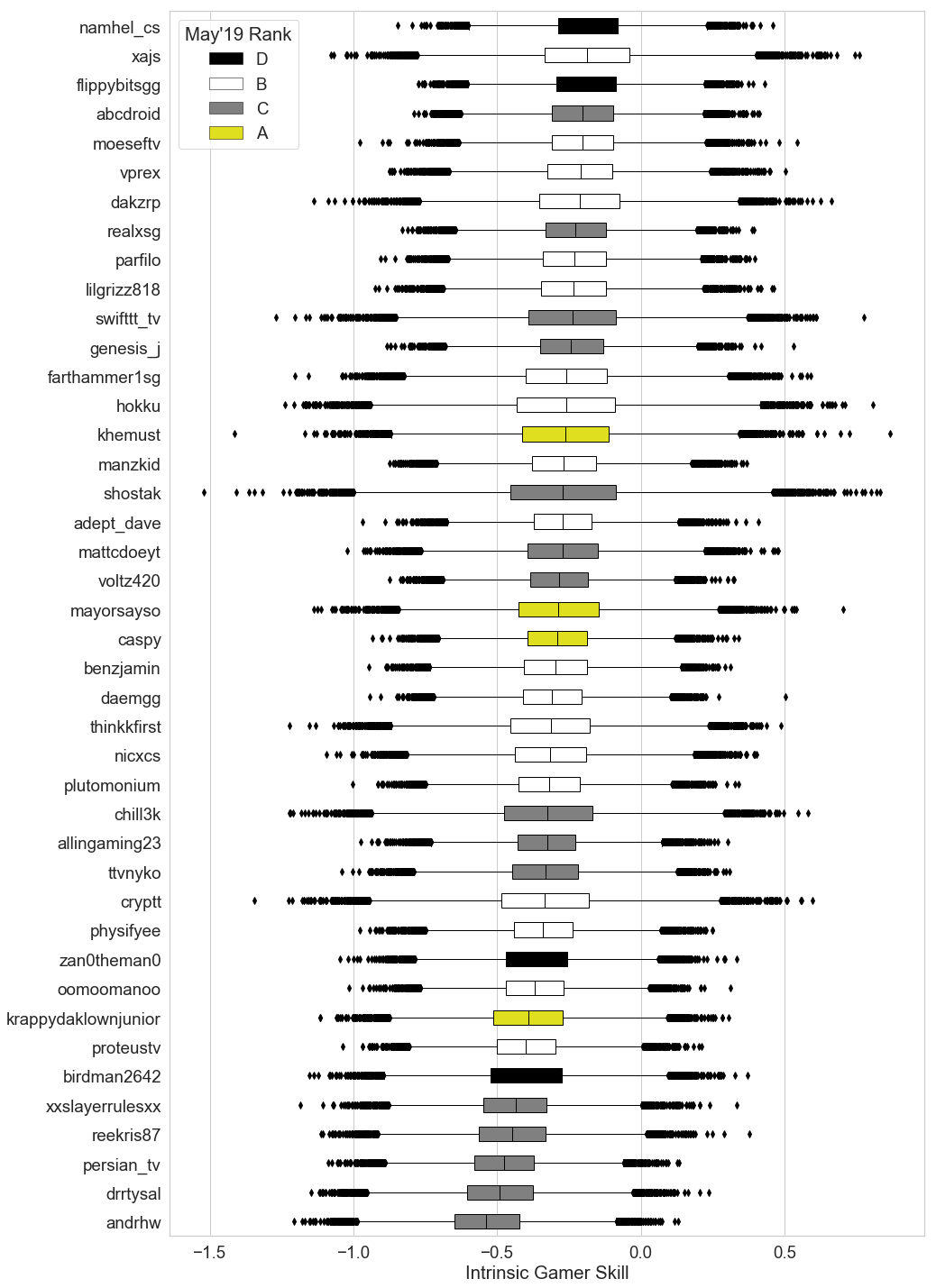}
    \caption{Gamers with the Bottom 10\% Jan'19-Apr'19 Data-based Intrinsic Skill Estimates and Their May'19 Ranks.}
    \label{fig:bottom_players}
\end{figure}

Figure~\ref{fig:top_players} and Figure~\ref{fig:bottom_players} show the extrinsic evaluation of our gamer-specific skill estimates against May 2019 ranks. Figure~\ref{fig:top_players} shows the top 10\% of the gamers (according to our estimates), while Figure~\ref{fig:bottom_players} shows bottom 10\% of the gamers. Each gamer is represented by a horizontal box plot that summarizes his or her posterior skill density. Notably, gamers with more abundant observational data have tighter credible intervals. The color labels correspond to the May 2019 rank. Across the figures, we can see that the top of our inter-gamer distribution is dominated by the gamers who are predominantly rank A section, with a few moving on to ranks G and S. On the other hand, the bottom of our inter-gamer distribution has only a handful of rank A section gamers.  

\begin{figure}[h]
    \centering
    \includegraphics[width=.45\textwidth]{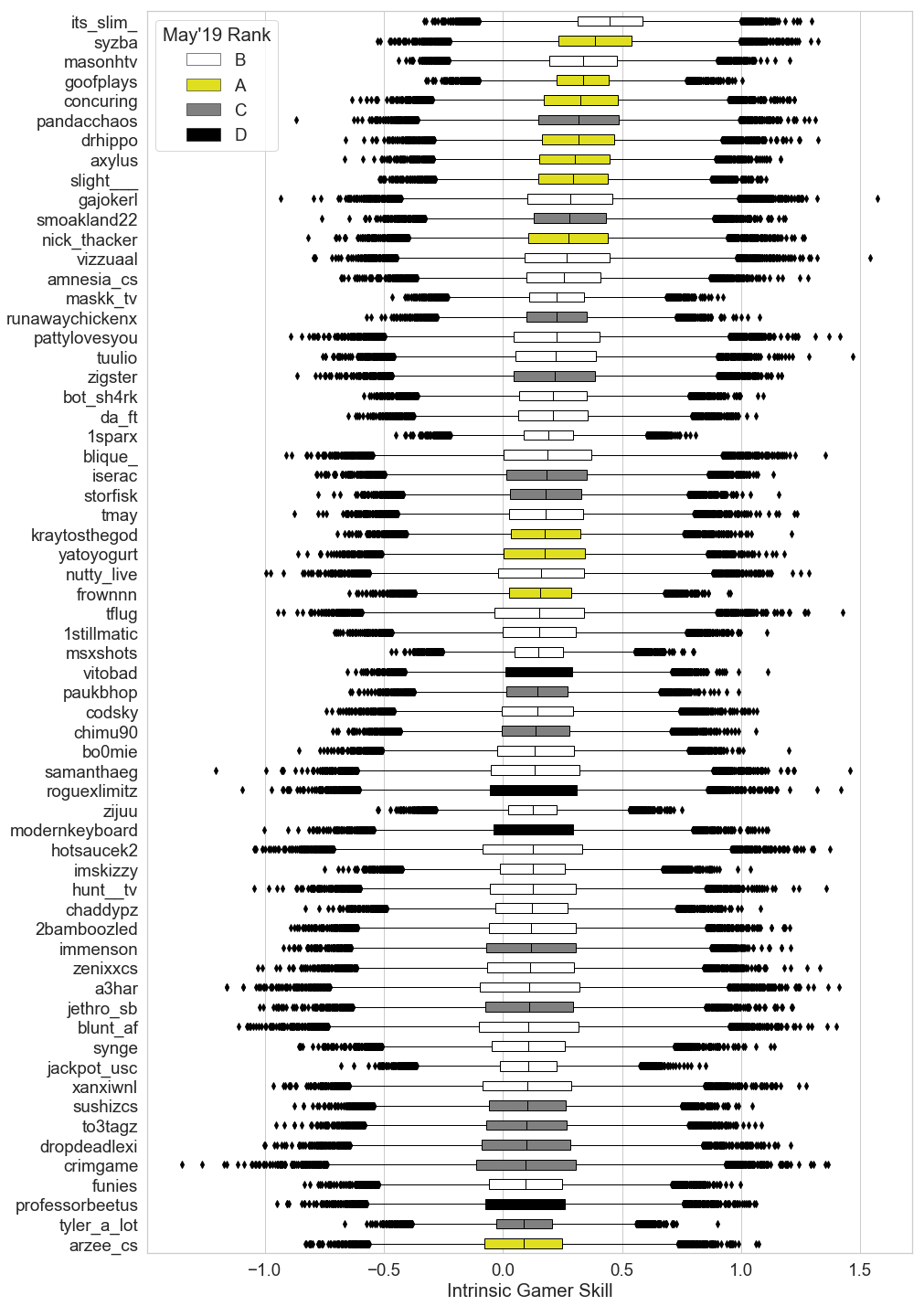}
    \caption{Low-Rank Gamers with Unusually High Jan'19-Apr'19 Data-based Intrinsic Skill Estimates and Their May'19 Ranks.}
    \label{fig:top_underA_players}
\end{figure}

Figure~\ref{fig:top_underA_players} shows the gamers who are ranked below A in our data, yet for whom we have generated relatively high skill predictions. Looking at the May 2019 ranks of these gamers, we can see quite a few conversions to rank A, with more of these conversions occurring at the top of this distribution. This pattern is statistically significant (Kendall $\tau$=0.297, p-value=0.00447). 
Finally, we have also tested whether conversions to rank G or S at the top of inter-gamer skill distribution (shown in Figure~\ref{fig:top_players}) are systematically correlated with the predicted skill, but did not find a statistically significant pattern.   

\section{Discussion}\label{sec:discussion}


Modeling and evaluation results that we report in Section~\ref{sec:results} imply the task of automated e-sports talent scouting based on Twitch chat and video stream data is likely a feasible one. Despite the fact that a gamer's leaderboard rank is affected by a plethora of factors that have not been represented in our modeling, such as teammate performance and equipment quality,\footnote{Note that this information is available for some of the ESEA profiled gamers. However, it is unlikely that it would be present in an actual e-sports scouting scenario involving unranked gamers.} we have been able to produce better-than-random predictions. Our models also have high recall, which may be important for capturing a larger talent pool for the subsequent human evaluation. Furthermore, our estimates of the intrinsic gamer skill---as opposed to his or her current leaderboard rank that may be obsolete or subject to randomness extrinsic to the gamer---are statistically significantly correlated with the improvements in gamers' leaderboard standings. 

We have found that the triplet loss produces the best results on all feature sets. However, pretraining is not uniformly helpful (e.g., models using video features achieve better performance if trained from scratch). We have also found that there are differences across the explored feature sets in terms of their effectiveness as predictors. Stronger features sets include temporal chat features (best AUC, F1, and precision) and motion video features (best recall). 
Performance of the temporal chat features is likely indicative of a strong relationship between the popularity of the player and the players' gaming skill.
In turn, our qualitative assessment of the visual features suggests that their effectiveness is likely tied to the specific visual elements that appear and persist on the screen whenever the player is in the ``time-out'' phase. 

In physical sports, player agility is an important indicator of skill; yet, we have found that the motion visual features are not uniformly better than the spatial visual features. It is possible that the variational motion estimation methods that we have used to estimate optical flow are not as good as the latest, neural methods \cite{ilg2017flownet}.  However, it is also possible that success of a gamer is related to more context-dependent, episodic moves (e.g., attacking from an ambush) that could have been missed either at the data collection (we record at most 5 minutes of gaming per day) or at the video processing stage (e.g., we aggressively downsample the videos as well as degrade image quality). We believe that further work on the motion visual feature representation could be beneficial. Furthermore, additional video data cleaning would undoubtedly be helpful, because there is no guarantee that the gamer is actually broadcasting the game, even though Twitch indicates that this is a CS:GO channel.\footnote{We have noted broadcasts of irrelevant activity, advertisements, as well as gamers waiting for his or her teammates to start the game.}

Feature sets that had not performed as well as we have hoped include audio features and chat content-based features. 
For the audio features, we have discovered early on that transcripts are not usable due to the multi-lingual nature of the Twitch platform\footnote{A quarter of the videos are not in English, judging from the title of the stream.} and relied only on the MFCC features. We found these features to be weak, possibly due to excessive quality degradation caused by downsampling (8K from 22.5K); with additional computational resources improvements of the MFCC feature quality could be explored. However, we note that gamers do different things during their gaming broadcasts: Some play music, some narrate their own game, some chat with teammates. Given the diversity and size of the audio data, we expect it to be quite noisy. Therefore, one of the improvements in audio feature extraction could be to focus on detection of specific sounds or tonal features from the audio track (e.g., yelling as an indicator of frustration), for example using SoundNet \cite{NIPS2016_6146}. 

The result for the chat content-based features is hardly surprising because this is an extremely noisy information channel. Twitch chats contain plenty of irrelevant information, e.g. hyperlinks for YouTube videos, emojis, nonsense characters.  Viewers repeat the same utterance multiple times for emotional emphasis as well as use many self-created game-specific words. Additionally, a significant portion of the chat content is non-English, despite our explicitly request for the stream language to be English at the data collection stage. All of these Twitch chat properties make it difficult for a traditional language classifier to effectively classify the chats. To make the textual chat features more effective, it would be helpful to implement additional data cleaning such that those obvious noises can be removed. Also, standard n-gram extractors do not work well due to the uncommon language structure used by the game viewers. Therefore, another improvement could be exploring other language representations that specifically work for the Twitch language.

Other limitations of our work are related to the relatively small number of unique players with ranking data (under 500). As such, we have been able to set aside fewer than 100 unique players for validation of the first-stage, modality-specific modeling. This has not been sufficient for application of discriminative learning techniques to develop a fusion model for modality-specific predictions. Instead, we relied on the Bayesian pooling approach that integrated the available evidence, yet did not explicitly model interactions across feature sets. A productive extension of our work could include development of a model that learns modality representations jointly, at the first modeling stage. 

\section{Conclusions}

We have investigated feasibility of a novel task of e-sports talent scouting based on multimodal Twitch chat and video stream data. In that, we have focused on CS:GO games broadcasted on Twitch during January 2019--April 2019. We have built two novel Twitch stream collections: One for 425 publicly ranked CS:GO gamers and one for 9,928 unranked CS:GO gamers. We have also developed a multimodal baseline model to make inferences about the intrinsic skills of the gamers. Our modeling results have been validated against May 2019 gamer ranks. In that we found that our intrinsic skill estimates are statistically significantly correlated with the future rank improvements. From this, we conclude that the automated e-sports talent scouting may, indeed, be a feasible task.

Researchers wishing to build on our work could investigate the benefits of improving Twitch data and/or representation quality, including development tied multimodal representations in the first modeling stage, or focus validating our conclusions using another online game represented on Twitch. 

\bibliography{main}

\begin{thebibliography}{49}
\providecommand{\natexlab}[1]{#1}
\providecommand{\url}[1]{\texttt{#1}}
\expandafter\ifx\csname urlstyle\endcsname\relax
  \providecommand{\doi}[1]{doi: #1}\else
  \providecommand{\doi}{doi: \begingroup \urlstyle{rm}\Url}\fi

\bibitem[Aslam \& Curry(2018)Aslam and Curry]{aslam2018towards}
Aslam, A. and Curry, E.
\newblock Towards a generalized approach for deep neural network based event
  processing for the internet of multimedia things.
\newblock \emph{IEEE Access}, 6:\penalty0 25573--25587, 2018.

\bibitem[Aytar et~al.(2016)Aytar, Vondrick, and Torralba]{NIPS2016_6146}
Aytar, Y., Vondrick, C., and Torralba, A.
\newblock Soundnet: Learning sound representations from unlabeled video.
\newblock In Lee, D.~D., Sugiyama, M., Luxburg, U.~V., Guyon, I., and Garnett,
  R. (eds.), \emph{Advances in Neural Information Processing Systems 29}, pp.\
  892--900. Curran Associates, Inc., 2016.

\bibitem[Bilal \& Erbad(2017)Bilal and Erbad]{bilal2017impact}
Bilal, K. and Erbad, A.
\newblock Impact of multiple video representations in live streaming: A cost,
  bandwidth, and qoe analysis.
\newblock In \emph{2017 IEEE International Conference on Cloud Engineering
  (IC2E)}, pp.\  88--94. IEEE, 2017.

\bibitem[Borth et~al.(2013)Borth, Chen, Ji, and Chang]{borth2013sentibank}
Borth, D., Chen, T., Ji, R., and Chang, S.-F.
\newblock Sentibank: large-scale ontology and classifiers for detecting
  sentiment and emotions in visual content.
\newblock In \emph{Proceedings of the 21st ACM international conference on
  Multimedia}, pp.\  459--460. ACM, 2013.

\bibitem[Bradski(2000)]{opencv_library}
Bradski, G.
\newblock {The OpenCV Library}.
\newblock \emph{Dr. Dobb's Journal of Software Tools}, 2000.

\bibitem[Cai et~al.(2007)Cai, He, and Han]{cai2007spectral}
Cai, D., He, X., and Han, J.
\newblock Spectral regression: A unified approach for sparse subspace learning.
\newblock In \emph{Seventh IEEE international conference on data mining (ICDM
  2007)}, pp.\  73--82. IEEE, 2007.

\bibitem[Cai \& Xia(2015)Cai and Xia]{cai2015convolutional}
Cai, G. and Xia, B.
\newblock Convolutional neural networks for multimedia sentiment analysis.
\newblock In \emph{Natural Language Processing and Chinese Computing}, pp.\
  159--167. Springer, 2015.

\bibitem[Campos et~al.(2017)Campos, Jou, and Giro-i Nieto]{campos2017pixels}
Campos, V., Jou, B., and Giro-i Nieto, X.
\newblock From pixels to sentiment: Fine-tuning cnns for visual sentiment
  prediction.
\newblock \emph{Image and Vision Computing}, 65:\penalty0 15--22, 2017.

\bibitem[Carpenter et~al.(2017)Carpenter, Gelman, Hoffman, Lee, Goodrich,
  Betancourt, Brubaker, Guo, Li, and Riddell]{carpenter2017stan}
Carpenter, B., Gelman, A., Hoffman, M.~D., Lee, D., Goodrich, B., Betancourt,
  M., Brubaker, M., Guo, J., Li, P., and Riddell, A.
\newblock Stan: A probabilistic programming language.
\newblock \emph{Journal of statistical software}, 76\penalty0 (1), 2017.

\bibitem[Cumming et~al.(2017)Cumming, Lloyd, Oliver, Eisenmann, and
  Malina]{cumming2017bio}
Cumming, S.~P., Lloyd, R.~S., Oliver, J.~L., Eisenmann, J.~C., and Malina,
  R.~M.
\newblock Bio-banding in sport: Applications to competition, talent
  identification, and strength and conditioning of youth athletes.
\newblock \emph{Strength \& Conditioning Journal}, 39\penalty0 (2):\penalty0
  34--47, 2017.

\bibitem[Deng(2018)]{deng2018profiling}
Deng, J.
\newblock \emph{Profiling Large-scale Live Video Streaming and Distributed
  Applications}.
\newblock PhD thesis, Queen Mary University of London, 2018.

\bibitem[Deng et~al.(2015)Deng, Cuadrado, Tyson, and Uhlig]{deng2015behind}
Deng, J., Cuadrado, F., Tyson, G., and Uhlig, S.
\newblock Behind the game: Exploring the twitch streaming platform.
\newblock In \emph{2015 International Workshop on Network and Systems Support
  for Games (NetGames)}, pp.\  1--6. IEEE, 2015.

\bibitem[Ffmpeg(2010)]{FFMPEG2010}
Ffmpeg.
\newblock Ffmpeg, 2010.
\newblock URL \url{http://www.ffmpeg.org}.

\bibitem[Ford et~al.(2017)Ford, Gardner, Horgan, Liu, Nardi, Rickman,
  et~al.]{ford2017chat}
Ford, C., Gardner, D., Horgan, L.~E., Liu, C., Nardi, B., Rickman, J., et~al.
\newblock Chat speed op pogchamp: Practices of coherence in massive twitch
  chat.
\newblock In \emph{Proceedings of the 2017 CHI Conference Extended Abstracts on
  Human Factors in Computing Systems}, pp.\  858--871. ACM, 2017.

\bibitem[Fu et~al.(2017)Fu, Lee, Bansal, and Berg]{fu2017video}
Fu, C.-Y., Lee, J., Bansal, M., and Berg, A.~C.
\newblock Video highlight prediction using audience chat reactions.
\newblock \emph{arXiv preprint arXiv:1707.08559}, 2017.

\bibitem[Gelman et~al.(1992)Gelman, Rubin, et~al.]{gelman1992inference}
Gelman, A., Rubin, D.~B., et~al.
\newblock Inference from iterative simulation using multiple sequences.
\newblock \emph{Statistical science}, 7\penalty0 (4):\penalty0 457--472, 1992.

\bibitem[Glickman et~al.(2018)Glickman, McKenzie, Seering, Moeller, and
  Hammer]{glickman2018design}
Glickman, S., McKenzie, N., Seering, J., Moeller, R., and Hammer, J.
\newblock Design challenges for livestreamed audience participation games.
\newblock In \emph{Proceedings of the 2018 Annual Symposium on Computer-Human
  Interaction in Play}, pp.\  187--199. ACM, 2018.

\bibitem[Gros et~al.(2018)Gros, Hackenholt, Zawadzki, and
  Wanner]{gros2018interactions}
Gros, D., Hackenholt, A., Zawadzki, P., and Wanner, B.
\newblock Interactions of twitch users and their usage behavior.
\newblock In \emph{International Conference on Social Computing and Social
  Media}, pp.\  201--213. Springer, 2018.

\bibitem[Haque(2019)]{haque2019twitch}
Haque, A.
\newblock Twitch plays pokemon, machine learns twitch: unsupervised
  context-aware anomaly detection for identifying trolls in streaming data.
\newblock \emph{arXiv preprint arXiv:1902.06208}, 2019.

\bibitem[He et~al.(2016)He, Zhang, Ren, and Sun]{he2016deep}
He, K., Zhang, X., Ren, S., and Sun, J.
\newblock Deep residual learning for image recognition.
\newblock In \emph{Proceedings of the IEEE conference on computer vision and
  pattern recognition}, pp.\  770--778, 2016.

\bibitem[He et~al.(2018)He, Gao, Xiao, Liu, and He]{he2018multimedia}
He, Z., Gao, S., Xiao, L., Liu, D., and He, H.
\newblock Multimedia data modelling using multidimensional recurrent neural
  networks.
\newblock \emph{Symmetry}, 10\penalty0 (9):\penalty0 370, 2018.

\bibitem[Hermans et~al.(2017)Hermans, Beyer, and Leibe]{hermans2017defense}
Hermans, A., Beyer, L., and Leibe, B.
\newblock In defense of the triplet loss for person re-identification.
\newblock \emph{arXiv preprint arXiv:1703.07737}, 2017.

\bibitem[Ilg et~al.(2017)Ilg, Mayer, Saikia, Keuper, Dosovitskiy, and
  Brox]{ilg2017flownet}
Ilg, E., Mayer, N., Saikia, T., Keuper, M., Dosovitskiy, A., and Brox, T.
\newblock Flownet 2.0: Evolution of optical flow estimation with deep networks.
\newblock In \emph{Proceedings of the IEEE conference on computer vision and
  pattern recognition}, pp.\  2462--2470, 2017.

\bibitem[Jiang et~al.(2019)Jiang, Qu, Wang, Wang, and Zheng]{jianglightor}
Jiang, R., Qu, C., Wang, J., Wang, C., and Zheng, Y.
\newblock Lightor: Video highlight detection for live streaming platforms using
  implicit crowdsourcing.
\newblock 2019.

\bibitem[Kendall(1938)]{kendall1938new}
Kendall, M.~G.
\newblock A new measure of rank correlation.
\newblock \emph{Biometrika}, 30\penalty0 (1/2):\penalty0 81--93, 1938.

\bibitem[Kung \& Hwang(1998)Kung and Hwang]{kung1998neural}
Kung, S.-Y. and Hwang, J.-N.
\newblock Neural networks for intelligent multimedia processing.
\newblock \emph{Proceedings of the IEEE}, 86\penalty0 (6):\penalty0 1244--1272,
  1998.

\bibitem[Laterman et~al.(2017)Laterman, Arlitt, and
  Williamson]{laterman2017campus}
Laterman, M., Arlitt, M., and Williamson, C.
\newblock A campus-level view of netflix and twitch: Characterization and
  performance implications.
\newblock In \emph{2017 International Symposium on Performance Evaluation of
  Computer and Telecommunication Systems (SPECTS)}, pp.\  1--8. IEEE, 2017.

\bibitem[Liu et~al.(2015)Liu, Lin, and Huang]{liu2015live}
Liu, Y.-W., Lin, C.-Y., and Huang, J.-L.
\newblock Live streaming channel recommendation using hits algorithm.
\newblock In \emph{2015 IEEE International Conference on Consumer
  Electronics-Taiwan}, pp.\  118--119. IEEE, 2015.

\bibitem[Ma et~al.(2018{\natexlab{a}})Ma, Cao, Li, and Chin]{ma2018tree}
Ma, W., Cao, K., Li, X., and Chin, P.
\newblock Tree structured multimedia signal modeling.
\newblock In \emph{The Thirty-First International Flairs Conference},
  2018{\natexlab{a}}.

\bibitem[Ma et~al.(2018{\natexlab{b}})Ma, Chang, Xu, Sebe, and
  Hauptmann]{ma2018joint}
Ma, Z., Chang, X., Xu, Z., Sebe, N., and Hauptmann, A.~G.
\newblock Joint attributes and event analysis for multimedia event detection.
\newblock \emph{IEEE transactions on neural networks and learning systems},
  29\penalty0 (7):\penalty0 2921--2930, 2018{\natexlab{b}}.

\bibitem[Marcel \& Rodriguez(2010)Marcel and
  Rodriguez]{Marcel:2010:TMP:1873951.1874254}
Marcel, S. and Rodriguez, Y.
\newblock Torchvision the machine-vision package of torch.
\newblock In \emph{Proceedings of the 18th ACM International Conference on
  Multimedia}, MM '10, pp.\  1485--1488, New York, NY, USA, 2010. ACM.
\newblock ISBN 978-1-60558-933-6.
\newblock \doi{10.1145/1873951.1874254}.
\newblock URL \url{http://doi.acm.org/10.1145/1873951.1874254}.

\bibitem[Miech et~al.(2018)Miech, Laptev, and Sivic]{miech2018learning}
Miech, A., Laptev, I., and Sivic, J.
\newblock Learning a text-video embedding from incomplete and heterogeneous
  data.
\newblock \emph{arXiv preprint arXiv:1804.02516}, 2018.

\bibitem[Nakandala et~al.(2017)Nakandala, Ciampaglia, Su, and
  Ahn]{nakandala2017gendered}
Nakandala, S.~C., Ciampaglia, G.~L., Su, N.~M., and Ahn, Y.-Y.
\newblock Gendered conversation in a social game-streaming platform.
\newblock In \emph{Eleventh International AAAI Conference on Web and Social
  Media}, 2017.

\bibitem[Platt et~al.(1999)]{platt1999probabilistic}
Platt, J. et~al.
\newblock Probabilistic outputs for support vector machines and comparisons to
  regularized likelihood methods.
\newblock \emph{Advances in large margin classifiers}, 10\penalty0
  (3):\penalty0 61--74, 1999.

\bibitem[Provensi et~al.(2017)Provensi, Eliassen, and
  Vitenberg]{provensi2017cloud}
Provensi, L., Eliassen, F., and Vitenberg, R.
\newblock A cloud-assisted tree-based p2p system for low latency streaming.
\newblock In \emph{2017 International Conference on Cloud and Autonomic
  Computing (ICCAC)}, pp.\  172--183. IEEE, 2017.

\bibitem[Raath(2017)]{raath2017rise}
Raath, A.
\newblock \emph{The rise of E-sports live-streaming and the resultant
  sponsorship opportunities presented to marketers}.
\newblock PhD thesis, The IIE, 2017.

\bibitem[Radicchi \& Mozzachiodi(2016)Radicchi and
  Mozzachiodi]{radicchi2016social}
Radicchi, E. and Mozzachiodi, M.
\newblock Social talent scouting: a new opportunity for the identification of
  football players?
\newblock \emph{Physical Culture and Sport. Studies and Research}, 70\penalty0
  (1):\penalty0 28--43, 2016.

\bibitem[Ruvalcaba et~al.(2018)Ruvalcaba, Shulze, Kim, Berzenski, and
  Otten]{ruvalcaba2018women}
Ruvalcaba, O., Shulze, J., Kim, A., Berzenski, S.~R., and Otten, M.~P.
\newblock Women’s experiences in esports: Gendered differences in peer and
  spectator feedback during competitive video game play.
\newblock \emph{Journal of Sport and Social Issues}, 42\penalty0 (4):\penalty0
  295--311, 2018.

\bibitem[Schumaker et~al.(2010)Schumaker, Solieman, and
  Chen]{schumaker2010sports}
Schumaker, R.~P., Solieman, O.~K., and Chen, H.
\newblock Sports knowledge management and data mining.
\newblock \emph{ARIST}, 44\penalty0 (1):\penalty0 115--157, 2010.

\bibitem[Simonyan \& Zisserman(2014)Simonyan and Zisserman]{simonyan2014two}
Simonyan, K. and Zisserman, A.
\newblock Two-stream convolutional networks for action recognition in videos.
\newblock In \emph{Advances in neural information processing systems}, pp.\
  568--576, 2014.

\bibitem[Sj{\"o}blom et~al.(2019)Sj{\"o}blom, T{\"o}rh{\"o}nen, Hamari, and
  Macey]{sjoblom2019ingredients}
Sj{\"o}blom, M., T{\"o}rh{\"o}nen, M., Hamari, J., and Macey, J.
\newblock The ingredients of twitch streaming: Affordances of game streams.
\newblock \emph{Computers in Human Behavior}, 92:\penalty0 20--28, 2019.

\bibitem[Taylor(2018)]{taylor2018watch}
Taylor, T.
\newblock \emph{Watch me play: Twitch and the rise of game live streaming}.
\newblock Princeton University Press, 2018.

\bibitem[Thelwall et~al.(2012)Thelwall, Buckley, and
  Paltoglou]{thelwall2012sentiment}
Thelwall, M., Buckley, K., and Paltoglou, G.
\newblock Sentiment strength detection for the social web.
\newblock \emph{Journal of the American Society for Information Science and
  Technology}, 63\penalty0 (1):\penalty0 163--173, 2012.

\bibitem[Wang et~al.(2007)Wang, Kim, Park, Lee, and Han]{wang2007learning}
Wang, D., Kim, Y.-S., Park, S.~C., Lee, C.~S., and Han, Y.~K.
\newblock Learning based neural similarity metrics for multimedia data mining.
\newblock \emph{Soft Computing}, 11\penalty0 (4):\penalty0 335--340, 2007.

\bibitem[Wang et~al.(2018)Wang, Ma, Jin, Yuan, Xun, Jha, Su, and
  Gao]{wang2018eann}
Wang, Y., Ma, F., Jin, Z., Yuan, Y., Xun, G., Jha, K., Su, L., and Gao, J.
\newblock Eann: Event adversarial neural networks for multi-modal fake news
  detection.
\newblock In \emph{Proceedings of the 24th ACM SIGKDD International Conference
  on Knowledge Discovery \& Data Mining}, pp.\  849--857. ACM, 2018.

\bibitem[Wirman(2018)]{wirman2018discourse}
Wirman, H.
\newblock \emph{The Discourse of Online Live Streaming on Twitch: Communication
  between Conversation and Commentary}.
\newblock PhD thesis, The Hong Kong Polytechnic University, Hong Kong, 2018.

\bibitem[Zach et~al.(2007)Zach, Pock, and Bischof]{zach2007duality}
Zach, C., Pock, T., and Bischof, H.
\newblock A duality based approach for realtime tv-l 1 optical flow.
\newblock In \emph{Joint pattern recognition symposium}, pp.\  214--223.
  Springer, 2007.

\bibitem[Zhang et~al.(2017)Zhang, Liu, Ma, Sun, and Li]{zhang2017seeker}
Zhang, C., Liu, J., Ma, M., Sun, L., and Li, B.
\newblock Seeker: Topic-aware viewing pattern prediction in crowdsourced
  interactive live streaming.
\newblock In \emph{Proceedings of the 27th Workshop on Network and Operating
  Systems Support for Digital Audio and Video}, pp.\  25--30. ACM, 2017.

\bibitem[Zhu et~al.(2017)Zhu, Yang, and Dai]{zhu2017understanding}
Zhu, Z., Yang, Z., and Dai, Y.
\newblock Understanding the gift-sending interaction on live-streaming video
  websites.
\newblock In \emph{International Conference on Social Computing and Social
  Media}, pp.\  274--285. Springer, 2017.

\end{thebibliography}
\bibliographystyle{icml2019}

\end{document}